# Generalized XGBoost Method

Yang Guang

*yg2357@163.com*

# Generalized XGBoost Method

The XGBoost method has many advantages and is especially suitable for statistical analysis of big data, but its loss function is limited to convex functions. In many specific applications, a nonconvex loss function would be preferable. In this paper, I propose a generalized XGBoost method, which requires weaker loss function constraint and involves more general loss functions, including convex loss functions and some non-convex loss functions. Furthermore, this generalized XGBoost method is extended to multivariate loss function to form a more generalized XGBoost method. This method is a multiobjective parameter regularized tree boosting method, which can model multiple parameters in most of the frequently-used parametric probability distributions to be fitted by predictor variables. Meanwhile, the related algorithms and some examples in non-life insurance pricing are given.

## 1.Introduction

The XGBoost library provides a regularized tree boosting method, as proposed by Chen and Guestrin (2016). This method has achieved excellent predictive performance in many fields and has exhibited many advantages, and is consequently considered especially suitable for the statistical analysis of big data. However, this method is limited because its loss function must be convex. For many scenario-specific problems, such as non-life insurance pricing, the distribution of predictor variables is often heavy-tailed, so the optimal prediction performance may not be obtained by setting convex loss functions. Simultaneously, it is important to estimate the probability distribution of predictor variables. When the set parametric probability distribution contains more than two parameters, it may be necessary to model multiple parameters to obtain better prediction performance. Therefore, a more generalized regularized tree boosting method is required to make the loss function not limited to the convex function while modelling the tree boosting for multiple parameters, to adapt to the most common parametric probability distributions.

Based on the XGBoost method, a generalized XGBoost method is proposed herein, which has weaker constraint for the loss function, which applies not only to convex loss functions. Furthermore, when the predictor variable is fitted with a multi-parameter probability distribution, it may be necessary to model multiple parameters of the probability distribution, so that the generalized XGBoost method is extended to the multivariate loss function to form a more generalized XGBoost method.

According to Duan et al. (2020), the NGBoost is also a method for multi-parameter tree boosting modelling of probability distributions. In contrast, the multiobjective parameter regularized tree boosting method proposed here has several major differences: the convergence of the algorithm is easily verified after meeting the constraint of the loss function. The regularization terms, especially those unique to XGBoost, are emphasized. The tree model is used to fit the increment $f_t(x_i)$ instead of gradient or natural gradient. Further, this method's principle is easy to understand for the general practitioner and is derived from the XGBoost method, which performs well in practice.

The remainder of this paper is organized as follows. Section 2 introduces the XGBoost method. Section 3 introduces the more generalized XGBoost method. Section 4 extends this generalized XGBoost method to multiobjective parameter and forms a multiobjective parameter regularized tree boosting method. Section 5 describes the proposed approach to combine the multiobjective parameter regularized tree boosting method and the multivariate loss function with the parameter estimation of the multi-parameter probability distribution. Section 6 provides examples of non-life insurance pricing.

## 2.XGBoost method

*XGBoost: A Scalable Tree Boosting System*

The following section was quoted from the paper of Chen, T. and Guestrin, C, *[XGBoost: A Scalable Tree Boosting System.](#)*

"For a given data set with *n* examples and *m* features $D = \{(x_i, y_i)\}(|D| = n, x_i \in R^m, y_i \in R)$, a tree ensemble model uses $K$ additive functions to predict the output.

$$\hat{y}_i = \emptyset(x_i) = \sum_{k=1}^{K} f_k(x_i), f_k \in F, \tag{1}$$

where $F = \{f(x) = \omega_{q(x)}\}(q: R^m \to T, \omega \epsilon R^T)$ was the space of regression trees (also known as CART or classification and regression tree). Here, $q$ represented the structure of each tree that mapped an example to the corresponding leaf index. $T$ was the number of leaves in the tree. Each $f_k$ corresponded to an independent tree structure $q$ and leaf weights $\omega$. Unlike decision trees, each regression tree contained a continuous score on each of the leaf, $\omega_i$ was used to represent the score on the i-th leaf. For example, the decision rules in the trees (given by $q$) were used to classify it into the leaves and calculate the final prediction by summing up the score in the corresponding leaves (given by $\omega$). To learn the set of functions used in the model, the following regularized objective was minimized.

$$L(\emptyset) = \sum_i l(\hat{y}_i, y_i) + \sum_k \Omega(f_k), \tag{2}$$

$$\text{where } \Omega(f_k) = \gamma T + \frac{1}{2}\lambda\|\omega\|^2.$$

Here, $l$ was a differentiable convex loss function that measured the difference between the prediction $\hat{y}_i$ and target $y_i$. The second term $\Omega$ penalized the complexity of the model (i.e., the regression tree functions). The additional regularization term helped to smooth the final learnt weights to avoid over-fitting. Intuitively, the regularized objective would tend to select a model employing simple and predictive functions. A similar regularization technique was used in Regularized greedy forest (RGF) model. Our objective and the corresponding learning algorithm was simpler than RGF and easier to parallelize. When the regularization parameter was set to zero, the objective fell back to the traditional gradient tree boosting.

The tree ensemble model in Equation (2) included functions as parameters and could not be optimized using traditional optimization methods in Euclidean space. Instead, the model was trained in an additive manner. Formally, let $\hat{y}_i^{(t)}$ be the prediction of the i-th instance at the t-th iteration, $f_t$ was added to minimize the following objective.

$$L^{(t)} = \sum_{i=1}^n l\left(y_i, \hat{y}_i^{(t-1)} + f_t(x_i)\right) + \Omega(f_t).$$

This greedily added the $f_t$ that most improved the model according to Equation (2). Second-order approximation was used to quickly optimize the objective in the general setting.

$$L^{(t)} \simeq \sum_{i=1}^n [l(y_i, \hat{y}_i^{(t-1)}) + g_i f_t(x_i) + \tfrac{1}{2} h_i f_t^2(x_i)] + \Omega(f_t)$$

where $g_i = \partial_{\hat{y}^{(t-1)}} l(y_i, \hat{y}_i^{(t-1)})$ and $h_i = \partial^2_{\hat{y}^{(t-1)}} l(y_i, \hat{y}_i^{(t-1)})$ were first and second order gradient statistics on the loss function. The constant terms were removed to obtain the following simplified objective at step $t$.

$$\tilde{L}^{(t)} = \sum_{i=1}^n \left[g_i f_t(x_i) + \tfrac{1}{2} h_i f_t^2(x_i)\right] + \Omega(f_t). \tag{3}$$

Define $I_j = \{i | q(x_i) = j\}$ as the instance set of leaf $j$. Equation (3) was rewritten by expanding $\Omega$ as follows:

$$\tilde{L}^{(t)} = \sum_{i=1}^n \left[g_i f_t(x_i) + \tfrac{1}{2} h_i f_t^2(x_i)\right] + \gamma T + \tfrac{1}{2} \lambda \sum_{j=1}^T \omega_j^2;$$

$$= \sum_{j=1}^{T}\left[\left(\sum_{i\in I_j} g_i\right)\omega_j + \frac{1}{2}\left(\sum_{i\in I_j} h_i + \lambda\right)\omega_j^2\right] + \gamma T. \tag{4}$$

For a fixed structure $q(x)$, the optimal weight $\omega_j^*$ of leaf $j$ was calculated by

$$\omega_j^* = -\frac{\sum_{i\in I_j} g_i}{\sum_{i\in I_j} h_i + \lambda}, \tag{5}$$

and the corresponding optimal value by

$$\tilde{L}^{(t)}(q) = -\frac{1}{2}\sum_{j=1}^{T}\frac{\left(\sum_{i\in I_j} g_i\right)^2}{\sum_{i\in I_j} h_i + \lambda} + \gamma T. \tag{6}$$

Equation (6) was used as a scoring function to measure the quality of a tree structure $q$. This score was similar to the impurity score for evaluating decision trees, except that it was derived for a wider range of objective functions.

Normally it is impossible to enumerate all the possible tree structures $q$; hence, a greedy algorithm that starts from a single leaf and iteratively adds branches to the tree is commonly used instead. Assume that $I_L$ and $I_R$ were the instance sets of left and right nodes after the split. Then, letting $I = I_L \cup I_R$ the loss reduction after the split be given as:

$$L_{split} = -\frac{1}{2}\left[\frac{\left(\sum_{i\in I_L} g_i\right)^2}{\sum_{i\in I_L} h_i + \lambda} + \frac{\left(\sum_{i\in I_R} g_i\right)^2}{\sum_{i\in I_R} h_i + \lambda} - \frac{\left(\sum_{i\in I} g_i\right)^2}{\sum_{i\in I} h_i + \lambda}\right] - \gamma. \tag{7}$$

This formula is often used in practice for evaluating the split candidates."

Moreover, the XGBoost method adopted shrinkage. "Shrinkage scales newly added weights by a factor $\eta$ after each step of tree boosting. Similar to a learning rate in tochastic optimization, shrinkage reduced the influence of each individual tree and left space for future trees to improve the model."

Other details of the XGBoost method can be found in the [original paper](original paper) and will not be covered here.

*Some shortcomings of XGBoost*

The XGBoost method required that the set loss function was a convex function. If it was not convex, the algorithm could not be guaranteed to converge to the global minimum. For instance, assume that there is only one sample point $(x_1, y_1)$ and the independent variable of $l(\hat{y}_1, y_1)$ is $\hat{y}_1$. The function plot is shown in Figure 1.

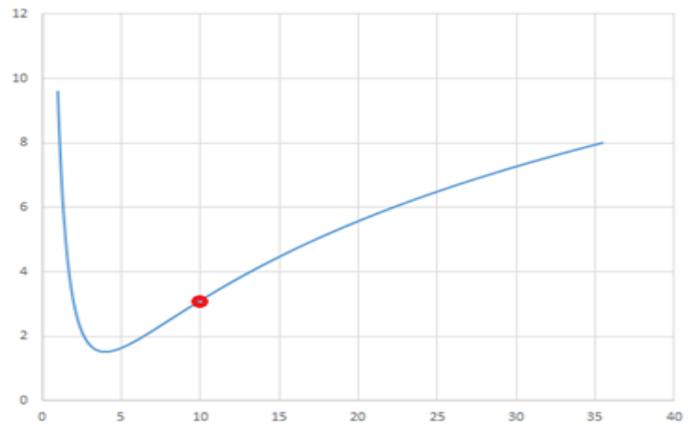

Figure 1. Example of a nonconvex loss function from example 1 in Section 6. Here, $l(\hat{y}_1, y_1)$ is a concave function in a neighbourhood of $\hat{y}_1 = 10$.

When γ and λ are quite small and negligible, the objective function is almost equivalent to the loss function. Testing the objective function can be avoided by testing the loss function instead, without affecting the conclusion.

Because there is only one sample point, $T = 1$. It is possible that the learning rate $\eta$ may not be controlled so that after $t - 1$-th iteration, $\hat{y}_1^{(t-1)} = 10$.

$l(\hat{y}_1, y_1)$ is a concave function in a neighborhood of $\hat{y}_1 = 10$. The first derivative $g_1$ of $\hat{y}_1$ is positive, but the second derivative $h_1$ is negative.

We provide the t-th iteration of the sample, $\omega_1^* = -\frac{\sum_{i \in I_1} g_i}{\sum_{i \in I_1} h_i + \lambda} = -\frac{g_1}{h_1 + \lambda}$. When $\lambda < |h_1|$, $\omega_1^* > 0$.

Hence, $\hat{y}_1^{(t)} = \hat{y}_1^{(t-1)} + \eta\omega_1^* > \hat{y}_1^{(t-1)}$, which deviates more from the global minimum point of $l(\hat{y}_1, y_1)$. Further, in XGBoost, the value range within which the parameter was estimated, $\hat{y}_i$, was all real numbers. In some scenarios, the value range of $\hat{y}_i$ was set to an interval of R. In this case, some algorithm adjustments were required.

### 3. More generalized XGBoost method

The more generalized XGBoost method relaxed the condition for the loss function in the XGBoost method. For the loss function $l(\hat{y}_i, y_i)$ of any sample $(x_I, y_i)$, where $\hat{y}_i$ was an independent variable, the constraint for $l(\hat{y}_i, y_i)$ were as follows:

(1) It was second-order differentiable;

(2) It includes one and only one local minimum point;

And only at that point, the derivative was zero or $l(\hat{y}_i, y_i)$ was strictly monotonic.

The tree model structure was the same as the XGBoost method, but the approximate expansion of the objective function was slightly different. Objective function, $L^{(t)} = \sum_{i=1}^n l\left(y_i, \hat{y}_i^{(t-1)} + f_t(x_i)\right) + \Omega(f_t)$, used one of the following approximations.

$$L^{(t)} \simeq \sum_{i=1}^n [l(y_i, \hat{y}_i^{(t-1)}) + g_i f_t(x_i)] + \Omega(f_t), \qquad (8)$$

or

$$L^{(t)} \simeq \sum_{i=1}^n [l(y_i, \hat{y}_i^{(t-1)}) + g_i f_t(x_i) + \tfrac{1}{2}\max(0, h_i) f_t^2(x_i)] + \Omega(f_t), \qquad (9)$$

or weighted average of equation (1) and (2)

$$L^{(t)} \simeq \sum_{i=1}^n [l(y_i, \hat{y}_i^{(t-1)}) + g_i f_t(x_i) + a \cdot \max(0, h_i) f_t^2(x_i)] + \Omega(f_t), a\epsilon \left[0, \tfrac{1}{2}\right]. \quad (10)$$

Equations (8) and (9) are special cases of equation (10).

In some cases, the value range of $\hat{y}_i$ need not be all real numbers; rather, they could be an interval of R. In general, we can set this interval to be closed.

At the algorithm level, XGBoost is improved accordingly. For the approximate formula (10),

$$\tilde{L}^{(t)} = \sum_{i=1}^{n}[g_i f_t(\boldsymbol{x_i}) + a \cdot \max(0, h_i) f_t^2(\boldsymbol{x_i})] + \gamma T + \frac{1}{2}\lambda \sum_{j=1}^{T} \omega_j^2$$

$$= \sum_{j=1}^{T}\left[\left(\sum_{i \in I_j} g_i\right)\omega_j + \left(a \cdot \sum_{i \in I_j} \max(0, h_i) + \frac{1}{2}\lambda\right)\omega_j^2\right] + \gamma T, a \in \left[0, \frac{1}{2}\right].$$

For a fixed tree structure $q(\boldsymbol{x})$, obtained the $\omega_j$ value where the $\tilde{L}^{(t)}$ for partial derivative of every $\omega_j$ was equal to 0, and the optimal weight score of leaf node $j$ is given as

$$\omega_j^* = -\frac{\sum_{i \in I_j} g_i}{2a \cdot \sum_{i \in I_j} \max(0, h_i) + \lambda}.$$

The optimal objective function value is given as

$$\tilde{L}^{(t)}(q) = -\frac{1}{2}\sum_{j=1}^{T} \frac{\left(\sum_{i \in I_j} g_i\right)^2}{\sum_{i \in I_j} 2a \cdot \max(0, h_i) + \lambda} + \gamma T.$$

The tree structure $q$ was obtained by the greedy algorithm, and branches were added iteratively from a single leaf node.

Assume $I_L$ and $I_R$ represented the sample set of left and right nodes after the split, $I = I_L \cup I_R$.

The reduced value of the objective function after splitting is given as

$$L_{split} = -\frac{1}{2}\left[\frac{\left(\sum_{i \in I_L} g_i\right)^2}{\sum_{i \in I_L} 2a \cdot \max(0, h_i) + \lambda} + \frac{\left(\sum_{i \in I_R} g_i\right)^2}{\sum_{i \in I_R} 2a \cdot \max(0, h_i) + \lambda} - \frac{(\sum_{i \in I} g_i)^2}{\sum_{i \in I} 2a \cdot \max(0, h_i) + \lambda}\right] - \gamma.$$

This equation is used for evaluating the split candidates. Similarly, shrinkage scales newly added weights by a factor $\eta$ after each step of tree boosting. Note that for some special samples $(\boldsymbol{x_i}, y_i)$ and some special $\hat{y}^{(t-1)}$, $g_i$ might be extremely large or even infinity, which affected the convergence or convergence rate of the algorithm. Therefore, a specific algorithm improvement is proposed herein, as given below.

If $|g_i|$ was greater than a certain large positive number $M$,

$$let\ \tilde{g}_i = \begin{cases} M, when & g_i \geq M, \\ g_i, when & -M < g_i < M, \\ -M, when & g_i \leq -M, \end{cases}$$

substituting $\tilde{g}_i$ for $g_i$. If not specifically mentioned, $g_i$ is still used for representing $\tilde{g}_i$ in this paper.

In some practical application scenarios, the value range of $\hat{y}_i$ was a specified interval of R. When $\eta$ was set to a relatively large value, some $\hat{y}_i^{(t)}$ might exceed its value range. If this case occurred, it was required to set the η value or set the value of $f_t(x_i)$ so that $\hat{y}_i^{(t)}$ was at the boundary of its value range.

It could be proved that the objective function $L^{(t)}$ converged to its global minimum when $\eta$ was small enough. The other details of the more generalized XGBoost method mentioned above were the same as the XGBoost method, which this paper will not specify.

It was advised to use the maximum likelihood estimate of $\hat{y}_i$ as the initial iteration value $\hat{y}_i^{(0)}$. The number of training rounds was reduced compared to the initial value of 0 and minimized the possibility of triggering adjustments to the $g_i$ mentioned above or adjustments to $\hat{y}_i^{(t)}$ beyond its value range.

**4. Extend the more generalized XGBoost method to multiobjective parameter**

This section extends the more generalized XGBoost method to multiobjective parameter and forms multiobjective parameter regularized tree boosting method. A sample set $D = \{(x_i, y_i)\}(|D| = n, x_i \in R^m, y_i \in R)$, contained $m$ characteristics and $n$ samples.

For any sample point $(x_I, y_i)$, consider $l$- variable loss function $l(\theta_{1_i}, ..., \theta_{l_i}; y_i)$, $\theta_{1_i}, ..., \theta_{l_i}$ is the independent variables, the value range of each $\theta_{j_i}(j = 1,2, ..., l)$ is a subinterval of R. The constraint of the loss function $l(\theta_{1_i}, \ldots, \theta_{l_i}; y_i)$ for multiobjective parameter regularized tree boosting method is

(1) $l(\theta_{1_i}, \ldots, \theta_{l_i}; y_i)$ is second-order differentiable, and there was one and only one local minimum point.

(2) Selected any $\theta_{j_i}$ and test $l(\theta_{j_i})$:

(2.1) There was only one local minimum point in $l(\theta_{j_i})$,

(2.2) The partial derivative of $l(\theta_{j_i})$ with respect to $\theta_{j_i}$ was zero only at the local minimum point, or $l(\theta_{j_i})$ is strictly monotonic.

Consider $\theta_{1_i}, \ldots, \theta_{l_i}$ as $l$ parameters to be estimated in the multiobjective parameter regularized tree boosting model.

Add $K_j$ tree functions to obtain the predicted result of parameter $\theta_{j_i}(j = 1, \ldots, l)$ of $l(\theta_{1_i}, \ldots, \theta_{l_i}; y_i)$.

$$\hat{\theta}_{j,i} = \emptyset_j(x_i) = \sum_{k=1}^{K_j} f_{\theta_j,k}(x_i), f_{\theta_j,k} \in F,$$

As in the XGBoost method, $F = \{f(x) = \omega_{q(x)}\}(q: R^m \to T, \omega \in R^T)$ was the space of regression trees. q represents the structure of each tree, which mapped a sample to the corresponding leaf index. $T$ was the number of leaves in the tree. Each $f_{\theta_j,k}$ corresponds to an independent tree structure $q$ and leaf weights $\omega$. To study these tree functions in the model, minimize the following regularization objectives:

$$L(\emptyset_1, \ldots, \emptyset_l) = \sum_i l(\theta_{1_i}, \ldots, \theta_{l_i}; y_i) + \sum_{k_1} \Omega_{\theta_1}(f_{k_1}) + \cdots + \sum_{k_l} \Omega_{\theta_l}(f_{k_l}),$$

where $\Omega_{\theta_1}(f_{k_1}) = \gamma_{\theta_1} T_{\theta_1} + \frac{1}{2}\lambda_{\theta_1}\|\omega_{\theta_1}\|^2$,

…,

$\Omega_{\theta_l}(f_{k_l}) = \gamma_{\theta_l} T_{\theta_l} + \frac{1}{2}\lambda_{\theta_l}\|\omega_{\theta_l}\|^2$.

Here, $\gamma_{\theta_j}, \lambda_{\theta_j}$ were the regularization hyperparameters of $\emptyset_j$, $T_{\theta_j}$ was the number of the of the leaf nodes of $\emptyset_j$ in the corresponding iteration round.

Similar to the XGBoost method, when the $t-1$-th iteration was determined, one of the following approximations was adopted for the objective function

$$L^{(t)} = \sum_{i=1}^{n} l\left(\widehat{\theta_{1_i}}^{(t-1)} + f_{\theta_1}^{(t)}(x_i), \ldots, \widehat{\theta_{l_i}}^{(t-1)} + f_{\theta_l}^{(t)}(x_i); y_i\right) + \Omega_{\theta_1}\left(f_{\theta_1}^{(t)}\right) + \ldots + \Omega_{\theta_l}\left(f_{\theta_l}^{(t)}\right)$$

of $t$-th iteration:

$$L^{(t)} \simeq \sum_{i=1}^{n} \left[ l\left(\widehat{\theta_{1_i}}^{(t-1)}, \ldots, \widehat{\theta_{l_i}}^{(t-1)}; y_i\right) + g_i^{\theta_1} f_{\theta_1}^{(t)}(x_i) + \cdots + g_i^{\theta_l} f_{\theta_l}^{(t)}(x_i)\right] + \Omega_{\theta_1}\left(f_{\theta_1}^{(t)}\right) + \cdots + \Omega_{\theta_l}\left(f_{\theta_l}^{(t)}\right) \quad (11)$$

or

$$L^{(t)} \simeq \sum_{i=1}^{n} \left[ l\left(\widehat{\theta_{1_i}}^{(t-1)}, \ldots, \widehat{\theta_{l_i}}^{(t-1)}; y_i\right) + g_i^{\theta_1} f_{\theta_1}^{(t)}(x_i) + \frac{1}{2}\max(0, h_i^{\theta_1})\left(f_{\theta_1}^{(t)}(x_i)\right)^2 + \cdots + g_i^{\theta_l} f_{\theta_l}^{(t)}(x_i) + \frac{1}{2}\max(0, h_i^{\theta_l})\left(f_{\theta_l}^{(t)}(x_i)\right)^2\right] + \Omega_{\theta_1}\left(f_{\theta_1}^{(t)}\right) + \cdots + \Omega_{\theta_l}\left(f_{\theta_l}^{(t)}\right) \quad (12)$$

or

$$L^{(t)} \simeq \sum_{i=1}^{n} \left[ l\left(\widehat{\theta_{1_i}}^{(t-1)}, \ldots, \widehat{\theta_{l_i}}^{(t-1)}; y_i\right) + g_i^{\theta_1} f_{\theta_1}^{(t)}(x_i) + a_1 \cdot \max(0, h_i^{\theta_1})\left(f_{\theta_1}^{(t)}(x_i)\right)^2 + \cdots + g_i^{\theta_l} f_{\theta_l}^{(t)}(x_i) + a_l \cdot \max(0, h_i^{\theta_l})\left(f_{\theta_l}^{(t)}(x_i)\right)^2\right] + \Omega_{\theta_1}\left(f_{\theta_1}^{(t)}\right) + \cdots + \Omega_{\theta_l}\left(f_{\theta_l}^{(t)}\right),$$

$$a_1, \ldots, a_l \in \left[0, \frac{1}{2}\right] \quad (13)$$

Equations (11) and (12) were special cases of equation (13).

Here,

$g_i^{\theta_j}$ is the partial derivative of a loss function $l\left(\widehat{\theta_{1_i}}^{(t-1)}, \ldots, \widehat{\theta_{l_i}}^{(t-1)}; y_i\right)$ to $\widehat{\theta_{j_i}}^{(t-1)}$, $h_i^{\theta_j}$ is the second partial derivative of a loss function $l\left(\widehat{\theta_{1_i}}^{(t-1)}, \ldots, \widehat{\theta_{l_i}}^{(t-1)}; y_i\right)$ to $\widehat{\theta_{j_i}}^{(t-1)}$

A maximum of $l$ trees could be trained simultaneously in each round of training (i.e., $l$ parameters to be estimated could be trained simultaneously). Each tree corresponded to one parameter to be estimated and had its own independent hyperparameters. Similar to the more generalized XGBoost method, if $|g_i|$ was greater than a certain large positive number M,

$$\text{let } \widetilde{g_i^{\theta_j}} = \begin{cases} M_j, & \text{when } g_i^{\theta_j} \geq M_j; \\ g_i^{\theta_j}, & \text{when } -M_j < g_i^{\theta_j} < M_j; \\ -M_j, & \text{when } g_i^{\theta_j} \leq -M_j. \end{cases}$$

substituting $\widetilde{g_i^{\theta_j}}$ for $g_i^{\theta_j}$. If not specifically mentioned, $g_i^{\theta_j}$ is still used for representing $\widetilde{g_i^{\theta_j}}$ in this paper.

Consider each parameter $\theta_j$ independently. $f_{\theta_j}^{(t)}(\boldsymbol{x_i})$ has the same tree structure and function expression as the more generalized XGBoost method.

For each parameter $\theta_j$ to be estimated, there were independent learning rate $\eta_j$, independent number of training rounds $K_j$, and independent super parameter $M_j$. Under certain scenarios, the value range of $\theta_j$ was an interval of R. In general, we can set this interval to be closed. If $\eta_j$ was set to a relative large value, some $\widehat{\theta_{j_i}}^{(t)}$ might exceed its value range. If this happens, it was needed to set the $\eta_j$ value, or set the $f_{\theta_j}^{(t)}(\boldsymbol{x_i})$ values to make $\widehat{\theta_{j_i}}^{(t)}$ fall at the boundary of the interval.

The value range of parameters could be reasonably selected. In practice, the reasonable prediction results would probably not fall at the theoretical extreme boundary condition. In general, the value range was considered a closed interval, and its boundary point had a reasonable distance from the theoretical boundary point. The other details of the algorithm are the same as the XGBoost method, which will not be specified in this paper.

It was straightforward to prove that the objective function $L^{(t)}$ converged to its global minimum when the learning rate was small enough. As discussed in Section 3, consider using the maximum likelihood estimate of $\theta_{1_i}, \ldots, \theta_{l_i}$ as the initial value of the iteration, which was quite beneficial.

## 5. Multiobjective parameter regularized tree boosting method and parameter estimation of parametric probability distributions

Assume that the predictor variable followed a parametric probability distribution that contained multiple parameters. To improve the prediction performance, multiple parameters was modelled, and then the specific expression of the probability distribution function was obtained. The multiobjective parameter regularized tree boosting method was an ideal method, and the motivation behind its proposition was to meet this requirement. It was assumed that the predictor variables of each sample point followed a certain parametric probability distribution and were independent of each other (conditionally independent based on their respective characteristics and parameters). In general, the negative log-likelihood function of the distribution was adopted as the loss function, and the distribution parameters to be estimated were the independent variables of the loss function. Assuming the independence of the sample point, the loss function of the sample set; namely, the sum of the loss functions of the sample points in the sample set, was the negative log-likelihood function of the sample set. As long as the loss function of each sample point satisfied the constraint for the loss function of the multiobjective parameter regularized tree boosting method, this method was applied to model and predict each parameter of a parametric probability distribution. It could be verified that, for most common parametric probability distributions and their common parameterized form, the negative log-likelihood function that was used as the loss function satisfied the constraint condition for the loss function of the multiobjective parameter regularized tree boosting method.

Similar to the generalized linear model, for the parameter to be estimated, some transformations of the parameter could be made, or a certain link could be added, as long as the constraint condition for the loss function of the multivariate regularized tree boosting method was satisfied. Similar to the application of the generalized linear model in practice, different parameterizations or links increased the number of candidate models, and the selection of the winning model among many candidate models were beneficial to improve the prediction performance.

**6.A few examples of non-life insurance pricing**

Similar to the XGBoost method, the more generalized XGBoost method and the multiobjective parameter regularized tree boosting method could be widely used in various fields.

The following section provides some examples in the non-life insurance pricing field.

*Example 1:*

It is assumed that the loss severity of auto insurance policy was subject to the gamma distribution $Y_i$. Here, $Y_i$ is independent of each other (conditionally independent under its respective characteristics and parameters) according to the classic non-life insurance pricing model. Gamma distribution was a classic heavy tail distribution. Regarding loss severity, the gamma distribution fit better than the normal distribution in most cases. Other details of the actuarial model for non-life insurance pricing can be found in classic textbooks, including the work of Klugman et al.

A classical parametric form of the probability density function of the gamma distribution is given by

$$f(y;\beta,\alpha) = \frac{\beta^{-\alpha}}{\Gamma(\alpha)} y^{\alpha-1} e^{-\frac{1}{\beta}y}, \alpha > 0, \beta > 0. \tag{14}$$

*where the expectation* $\mu = \alpha \cdot \beta$, $\beta = \frac{\mu}{\alpha}$.

Similar to the generalized linear model, our interest focuses on the expectation of loss severity. Now the more generalized XGBoost method was applied to fit the μ.

We rewrote equation (14) as

$$f(y;\mu,\alpha) = \frac{(\frac{\alpha}{\mu})^\alpha}{\Gamma(\alpha)} y^{\alpha-1} e^{-\frac{\alpha}{\mu}y}, \text{ where } \mu = \alpha \cdot \beta.$$

The probability density function of a loss severity of the auto insurance policy is given as

$$f(y_i;\mu_i,\alpha) = \frac{(\frac{\alpha}{\mu_i})^\alpha}{\Gamma(\alpha)} y_i^{\alpha-1} e^{-\frac{\alpha}{\mu_i}y_i}.$$

As assumed by independence, the loss function of the training set may be expressed as

$$\sum_{i=1}^n l(\hat{y}_i; y_i, \alpha) = -\sum_{i=1}^n \ln f(\hat{y}_i; y_i, \alpha) = -\sum_{i=1}^n [\alpha\ln\alpha - \alpha\ln\hat{y}_i - \ln\Gamma(\alpha) + (\alpha-1)\ln y_i - \frac{\alpha}{\hat{y}_i} y_i], \text{ where } \hat{y}_i \text{ denotes } \hat{\mu}_i.$$

When $\alpha = 5$, $y_i = 4$, the plot of $l(\hat{y}_i; y_i, \alpha)$ is shown as that in Figure 2.

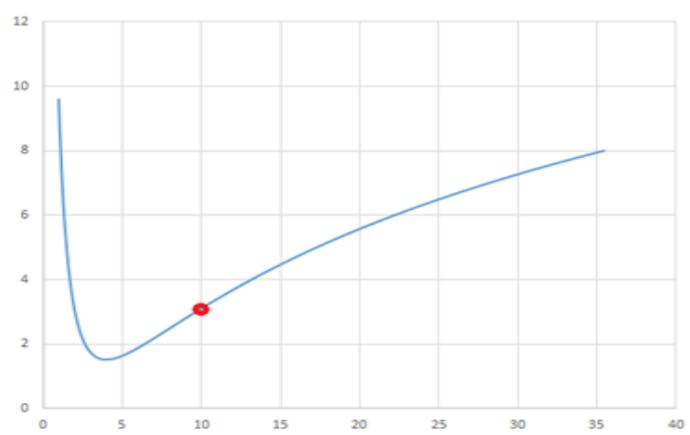

Figure 2. Loss function plot of a sample. The negative log likelihood function of gamma distribution was used as the loss function, where $\alpha = 5$ and $y_i = 4$.

It was not a convex function and did not satisfy the loss function requirement of the XGBoost method. It could be verified that for any $y_i$ and α, $l(\hat{y}_i; y_i, \alpha)$ was second-order differentiable to $\hat{y}_i$; there was one and only one local minimum point and only at that point, the derivative was zero. Thus, the loss function constraint condition of the more generalized XGBoost method was satisfied. The estimation parameter $\mu_i$ was modelled using the more generalized XGBoost method: $\hat{y}_i = \emptyset(x_i)$. Here, $\emptyset(x_i)$ was the tree function of the more generalized XGBoost method.

Consider $\alpha$ as a nuisance parameter. Its value could be determined empirically. It could also be considered as a hyperparameter, and its value was determined by other methods such as grid searching or maximum likelihood estimation .

*Example 2:*

It is assumed that the loss frequency of auto insurance policy was subject to the mixed distribution $Y_i$ of the degenerate zero distribution (one-point distribution) and Poisson distribution. The probability distribution function was

$$\begin{cases} P(Y=0) = (1-\alpha) + \alpha e^{-\lambda} \\ P(Y=k) = \alpha \frac{\lambda^k}{k!} e^{-\lambda}, \ k=1,2,\ldots \end{cases}, \ \alpha \in (0,1], \lambda \in (0, +\infty).$$

This distribution belonged to $(a, b, 1)$ class and did not belong to the exponential family, where $\mu = E(Y) = \alpha\lambda$. Assume that $Y_i$ was independent of each other (conditionally independent under their respective characteristics and parameters).

Similar to the generalized linear model, our interest focuses on the expectation of loss frequency. The probability distribution function of $Y_i$ can be written as

$$P(Y_i; \mu_i, \alpha) = \begin{cases} (1-\alpha) + \alpha e^{-\frac{\mu_i}{\alpha}}, & y_i = 0 \\ \alpha \frac{\left(\frac{\mu_i}{\alpha}\right)^{y_i}}{y_i!} e^{-\frac{\mu_i}{\alpha}}, & y_i = 1,2,\ldots \end{cases}, \ \alpha \in [0,1], \mu_i \in (0, +\infty).$$

Take the negative log-likelihood function as the loss function and take $\mu_i$ as the parameter to be estimated, and its predicted value was expressed as $\hat{y}_i$. Based on the independence hypothesis, the loss function of the whole training set was

$$\sum_{i=1}^n l(\hat{y}_i; y_i, \alpha) = \sum_{i=1}^n -lnP(y_i, \hat{y}_i, \alpha) \text{。}$$

When $\alpha = 0.5$ and $y_i = 0$, the plot of $l(\hat{y}_i; y_i, \alpha)$ is shown as that in Figure 3.

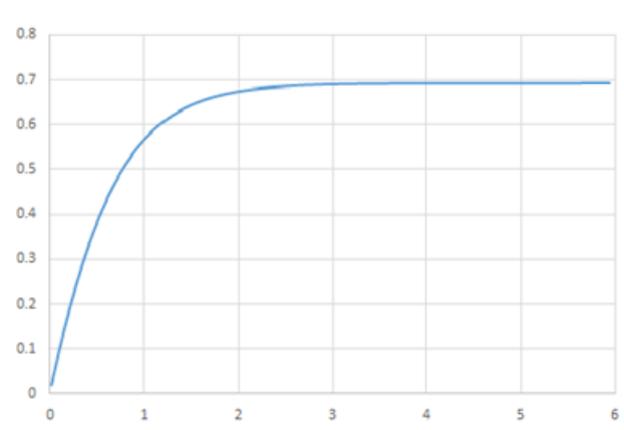

Figure 3. Loss function plot of a sample. The negative log likelihood function was used as the loss function, where $\alpha = 0.5$ and $y_i = 0$.

It was not a convex function and did not satisfy the loss function constraint of the XGBoost method. It could be verified that, for any $y_i$ and α, $l(\hat{y}_i; y_i, \alpha)$ was second-order differentiable to $\hat{y}_i$; there was one and only one local minimum point; and only at that point, the derivative was zero, or $l(\hat{y}_i; y_i, \alpha)$ was strictly monotonic. Thus, the loss function constraint of the more generalized XGBoost method was satisfied.

The estimation parameter $\mu_i$ was modelled using the more generalized XGBoost method: $\hat{y}_i = \emptyset(\boldsymbol{x}_i)$. $\emptyset(\boldsymbol{x}_i)$, which was the tree function of the more generalized XGBoost method. Consider $\alpha$ as a nuisance parameter.

Comment: This mixed distribution had a larger variance than the expectation and was often applied to fit over-dispersed data that have many zeros, which was one of the options for solving the problem of overdispersion caused by Poisson distribution fitting. Further, this distribution belonged to (a, b,1) class and did not belong to the exponential family. Thus, compared with GLM, the more generalized XGBoost method could fit a wider range of parametric probability distributions.

*Example 3:*

Compared with Poisson distribution, negative binomial distribution could also solve overdispersion in the fitting process. It is assumed that the loss frequency of auto insurance policy followed a negative binomial distribution $Y_i$, and a classical parameter form of the probability distribution function of this distribution was

$$P(Y_i; \beta_i, \gamma_i) = \binom{y_i + \gamma_i - 1}{y_i} \left(\frac{1}{1+\beta_i}\right)^{\gamma_i} \left(\frac{\beta_i}{1+\beta_i}\right)^{y_i}, y_i = 0,1,2,\ldots,$$

$$\beta_i > 0, \gamma_i > 0.$$

We were interested in both parameters of the negative binomial distribution. Taking the negative log-likelihood function as the loss function, and assuming its independence, the loss function of the training set was

$$\sum_{i=1}^{n} l(\theta_{1_i}, \theta_{2_i}; y_i) = \sum_{i=1}^{n} -lnP(y_i, \beta_i, \gamma_i)$$

It is easy to verify that for some $y_i$ and $\gamma_i$, the loss function $-lnP(y_i, \beta_i, \gamma_i)$ is not the convex function of $\beta_i$. The value range of the parameter could be set to $\beta_i \in [\varepsilon_1, M_1], \gamma_i \in [\varepsilon_2, M_2]$, where $\varepsilon_1, \varepsilon_2$ were small enough positive numbers, while $M_1, M_2$ were large enough positive numbers.

It could be verified that for any $y_i$, the loss function $-lnP(\beta_i, \gamma_i; y_i)$ met the constraint for loss function in the value range of the multiobjective parameter regularized tree boosting method.

Therefore, the parameter estimation of the probability distribution could be obtained by using the corresponding method.

Furthermore, considering that the sample points have different exposure numbers, similar to Poisson regression, exposure refers to the duration of the observation time or the size of the observed space or the number of micro individuals within a sample point, etc. For ordinary scenarios and models, exposure and the expectation of predictor variables are directly proportionate. We took one car per year as one exposure unit. Similar to the widely used Poisson regression's treatment of exposure number, it was assumed that the observed samples followed the negative binomial distribution of the following parameter form:

$$P(Y; \beta, \gamma, exposure) = \binom{y + exposure \cdot \gamma - 1}{y} \left(\frac{1}{1+\beta}\right)^{exposure \cdot \gamma} \left(\frac{\beta}{1+\beta}\right)^y,$$

where $exposure$ represents the exposure number of the sample.

Taking the negative log-likelihood function as the loss function, the loss function of a sample point $(x_i, y_i)$ was

$$l(\beta_i, \gamma_i; y_i, exposue_i) = -\left[\ln\binom{y_i + exposure_i \cdot \gamma_i - 1}{y_i} + exposure_i \cdot \gamma_i \cdot \ln\left(\frac{1}{1+\beta_i}\right) + y_i \ln\left(\frac{\beta_i}{1+\beta_i}\right)\right].$$

It can be proved for any $y_i$ and $exposue_i$, the constraint for the loss function of the multiobjective parameter regularized tree boosting method is satisfied in the value range. Moreover, different insurance policies might have different terms and conditions on deductibles for a claim, and the deductible amount determines whether to report a loss if it occurs. The underwriter obtains first-hand data about loss frequency that is actually a claim frequency. To customize different deductibles to the insured party, distributions of loss frequency are calculated backwards from the claim frequency distribution and deductibles as well as loss severity distribution. Here, the loss refers to the claim amount that might occur if the policy does not have deductibles. Under the classic non-life insurance pricing actuarial theories, similar to the exposure number, I adjusted the parameter $\beta$, where $a$ was the adjustment coefficient. It was assumed that the observed sample followed a negative binomial distribution in the following parameter form:

$$P(Y; \beta, \gamma, exposure, a) = \binom{y + exposure \cdot \gamma - 1}{y}\left(\frac{1}{1+a\cdot\beta}\right)^{exposure\cdot\gamma}\left(\frac{a\cdot\beta}{1+a\cdot\beta}\right)^y.$$

Take the negative logarithmic likelihood function as the loss function:

$$l(\beta_i, \gamma_i; y_i, exposue_i, a_i) = -\left[\ln\binom{y_i + exposure_i \cdot \gamma_i - 1}{y_i} + exposure_i \cdot \gamma_i \cdot \ln\left(\frac{1}{1+a_i\beta_i}\right) + y_i \ln\left(\frac{a_i\beta_i}{1+a_i\beta_i}\right)\right]。$$

It could be proved for any $y_i$, $a_i$, and $exposue_i$, the constraint for the loss function of the multiobjective parameter regularized tree boosting method was satisfied in the value range.

Comment: In non-life insurance pricing, the distribution of the predictor variable (not just the expectation or variance of the loss variable) is important for at least two reasons enumerated below:

1. The insurer needs to arrange reinsurance according to the probability distribution of the total claim variables to control the overall risk.

2. For different deductibles, different probability distributions of claim frequency needs to be determined to carry out customized pricing.

The above cases, which combined with the most advanced machine learning techniques and mature insurance actuary theories, might open up new ideas in the application of big data analysis in the non-life insurance actuarial field.

## 7.Discussion

The selection of the evaluation index had to be consistent with the selection of loss function. For the loss function which took the negative log-likelihood function as the training set, it was natural to take the negative log-likelihood function as the evaluation index for the verification set and the test set. In this way, different parameter forms, different candidate parametric probability distributions, and even different modelling methods including XGBoost method, more generalized XGBoost method, multiobjective parameter regularized tree boosting method, and generalized linear model etc., were compared under the same framework, and the winning model was selected by using the test set.
When the interest was focused on a certain parameter, and the remaining parameters were highly deterministic, a more generalized XGBoost method was applied to process the uninterested parameters as nuisance parameters. If multiple parameters were interested simultaneously, it was necessary to adopt the multiobjective parameter regularized tree boosting method. Similarly, if some parameters in the multi-parameter distribution were less interested or more deterministic, they could be also processed as nuisance parameters.

In the multiobjective parameter regularized tree boosting method, for the parameters to be estimated with strong certainty, its main regularization hyperparameters, such as $\gamma$ and $\lambda$, needed to be increased to prevent overfitting. To improve the calculation efficiency, the number of training rounds corresponding to these parameters were reduced. A better solution was to set the iterations intervals so that the total number of training rounds was reduced. The certainty of parameters to be estimated was based on Bayesian estimation theory. Moreover, the [Bühlmann Credibility Theory](#) could be combined in practice especially in actuarial pricing. In the multiobjective parameter regularized tree boosting method, different reasonable learning rates were set for different estimation parameters, which improved the convergence rate. This generalized approach to the XGBoost method mentioned here was also applicable to methods such as [LightGBM](#) and [CatBoost](#), which were improved based on the XGBoost method.